\documentclass[conference]{IEEEtran}
\usepackage{amsmath,amssymb,amsfonts}
\usepackage{array}
\usepackage[caption=false,font=normalsize,labelfont=sf,textfont=sf]{subfig}
\usepackage{textcomp}
\usepackage{stfloats}
\usepackage{url}
\usepackage{verbatim}
\usepackage{graphicx}
\usepackage{cite}
\usepackage{xcolor}
\usepackage{booktabs,siunitx}
\usepackage[ruled,vlined, linesnumbered]{algorithm2e}
\def\BibTeX{{\rm B\kern-.05em{\sc i\kern-.025em b}\kern-.08em
    T\kern-.1667em\lower.7ex\hbox{E}\kern-.125emX}}

\begin{document}

\title{Enabling Heterogeneous Domain Adaptation in Multi-inhabitants Smart Home Activity Learning}

\author{$^1$Md Mahmudur Rahman and $^2$Mahta Mousavi and $^3$Peri Tarr
 and $^{1,4}$Mohammad Arif Ul Alam\\
$_1$University of Massachusetts Lowell\\
$_2$University of California, San Diego\\
$_3$IBM's Thomas J. Watson Research Center\\
$_4$University of Massachusetts Chan Medical School\\
        \small{ MdMahmudur\_Rahman@student.uml.edu}
        \small{  mmousavi@eng.ucsd.edu}
        \small{  tarr@us.ibm.com}
        \small{  mohammadariful\_alam@uml.edu}
}



\maketitle

\begin{abstract}

Domain adaptation for sensor-based activity learning is of utmost importance in remote health monitoring research. However, many domain adaptation algorithms suffer with failure to operate adaptation in presence of target domain heterogeneity (which is always present in reality) and presence of multiple inhabitants dramatically hinders their generalizability producing unsatisfactory results for semi-supervised and unseen activity learning tasks. We propose \emph{AEDA}, a novel deep auto-encoder-based model to enable semi-supervised domain adaptation in the existence of target domain heterogeneity and how to incorporate it to empower heterogeneity to any homogeneous deep domain adaptation architecture for cross-domain activity learning. Experimental evaluation on 18 different heterogeneous and multi-inhabitants use-cases of 8 different domains created from 2 publicly available human activity datasets (wearable and ambient smart homes) shows that \emph{AEDA} outperforms (max. 12.8\% and 8.9\% improvements for ambient smart home and wearables) over existing domain adaptation techniques for both seen and unseen activity learning in a heterogeneous setting.
\end{abstract}


\section{Introduction}

Remote monitoring of elderly activities helps to manage chronic disease, post-acute care, and monitoring safety thus building sustainable healthcare models in smart environments. The advent of wearable and ambient devices i.e. Internet of Things (IoT) in conjunction with machine learning techniques help continuous monitoring of activities remotely. The key to the success of many existing activity recognition algorithms is the availability of abundant labeled training data \cite{lara2012survey}. However, for many real-world problems, collecting labeled activity data is often very expensive, cumbersome, time consuming and erroneous. As older adults are reluctant of all efficient ground-truth labeling methods such as using cameras, self-reporting activity logs or activity tagging by peers, activity labeling has become one of the major issues in remote activity monitoring research based on supervised or semi-supervised learning algorithms. On the other hand, due to the immense presence of different heterogeneity of sensor environment such as smart home structure, number of sensors, sensor firing sequence and location heterogeneity, even the most powerful to date domain adaptation technique also stagnates at poor accuracy (avg. 82\%) in practice for activity learning \cite{DBLP:journals/corr/SunS16a}.


\par Domain adaptation and transfer learning help to overcome scarcity of labeled data in target domain by utilizing information about the task, and data from single or multiple auxiliary domains (referred to as source domain). Recent advancement of deep learning encourages some successful supervised/unsupervised domain adaptation algorithms in the activity recognition domain \cite{tzeng2017adversarial} \cite{ganin2016domain}. 
There are several unsupervised activity domain adaptation frameworks based on hidden Markov model, graphical model or deep learning that provide significant improvement in a homogeneous environment \cite{wang18} \cite{wyatt2005unsupervised} \cite{attal2015physical}. However, nearly all of the previous models fail in addressing the heterogeneity in terms of home structure, sensor types, sensor location and number of inhabitants which are always present in real-world smart homes. To solve the problem, we propose \emph{AEDA}, an auto-encoder based semi-supervised domain adaptation method. Then, we propose a technique of enabling any homogeneous domain adaptation to work in a heterogeneity setup using the power of \emph{AEDA}. 
 \par 
 In \emph{AEDA}, we use one auto-encoder each on the source and target data to map both domains on a feature space of similar distribution with a novel loss function.
 As we use different auto-encoders as input network of source and target datasets, this architecture has become highly capable of handling heterogeneity of wide margin between source and target domain. The {\bf key contributions} are:
 {\it
 \begin{itemize}
    \item \emph{AEDA}, a novel ensemble sequence-to-sequence auto-encoder based Semi-supervised  domain  adaptation (SSDA) algorithm that improves the accuracy significantly in cross-home activity learning for both single and multiple inhabitants.
    \item A novel technique of using \emph{AEDA} to empower existing deep homogeneous domain adaptation techniques to work in heterogeneous target data in semi-supervised setup.
    \item Evaluate proposed frameworks on 12 different heterogeneous and multi-inhabitant use-cases by using two publicly available human activity data-sets (wearable and ambient smart homes) and show our frameworks outperform state-of-art algorithms.
 \end{itemize}
}
\section{Related Works}

As deep neural networks are good at capturing complex features, different deep learning-based domain adaptation methods are successful in a large variety of domains. Glorot et. al. proposed a Stacked Denoising Auto-encoder (SDA)-based domain adaptation method for sentiment analysis \cite{glorot2011domain} and a residual transfer network was proposed in \cite{long2016unsupervised}. The adversarial network-based approach used a discriminator network to map source and target feature spaces to a feature space of similar distribution \cite{tzeng2017adversarial}. Similarly, in \cite{ganin2016domain}, a domain classifier with a gradient reversal layer was used for common feature space mapping. In other auto-encoder based approaches, marginalized auto-encoder was used to learn common features \cite{chen2012marginalized} and reconstruction of source domain samples from target domain samples was used with bi-shifting auto-encoder\cite{kan2015bi}.

While, the main challenge in deep learning based domain adaptation is the gap in feature distributions between domains, which degrades the source classifier's performance, the recent works have been focusing on unsupervised domain adaptation (UDA) and, in particular, feature distribution alignment. Interestingly, we empirically observe that UDA methods \cite{ganin2016domain,ming18,kuniaki18} often failed in improving accuracy in SSDA i.e. when a small fraction of labeled data is available in the target dataset. Daume \textit{et al.}\cite{daume2010frustratingly} proposed the semi-supervised version of the supervised EasyAdapt (EA) algorithm which uses augmented feature space to map between source and target domains. On the other hand, feature space independent kernel matching method was proposed in \cite{xiao2014feature}. In \cite{ao2017fast}, authors used a completely different `soft label' approach to enable the target model to mimic the output of the source model.

For heterogeneous domain adaptation, different approaches has been proposed, including augmented feature space \cite{li2013learning}, manifold alignment \cite{wang2011heterogeneous}, cross-domain landmark \cite{hubert2016learning} and discriminative correlation subspace \cite{yan2017learning} and relative distribution of network weights \cite{khan2018scaling} based approaches, but, very few researchers considered the presence of heterogeneity between source and target datasets. In \cite{li2013learning}, authors used augmented representations of heterogeneous features to learn the common features and Wang \textit{et al.} proposed a manifold alignment method where labeled data from multiple sources are reused \cite{wang2011heterogeneous}. However,  Hubert \textit{et al.} used cross-domain landmark selection to derive a domain invariant feature subspace \cite{hubert2016learning}. Similarly, a common subspace-based method was proposed in \cite{yan2017learning} where a correlation subspace is mapped discriminately. Finally, Khan \textit{et al.} used the relative distribution of corresponding Convolutional Neural Network (CNN) layers to map in a common feature space \cite{khan2018scaling}. However, very few of the aforementioned research works ever considered the heterogeneous semi-supervised deep domain adaptation in any domain. In this paper, we design and develop an auto-encoder based \emph{AEDA} model to activate semi-supervised domain adaptation in diverse set of sensors (heterogeneous)-assisted smart home use-cases for cross-domain activity learning

\section{Problem Formulation}

To formulate the semi-supervised domain adaptation problem, we consider an activity source domain $\mathcal{D}^s$ which consists of a sensor space (feature space) $\mathcal{X}^s$ with dimension $d^s$ and a marginal probability distribution $P(X^s)$, where $X^s=\{x_1,\ldots,x_{n_s}\}\in \mathcal{X}^s$ and $n_s$ is number of sensors. Given a specific source domain $\mathcal{D}^s =\{\mathcal{X}^s,P(X)^s\}$, source activity classification task $\mathcal{T}^s$ consists of an activity label space $\mathcal{Y}^s$ and an objective predictive function $f(\cdot)$, which can also be viewed as a conditional probability distribution $P(Y^s|X^s)$ from a probabilistic perspective, where $X^s \in \mathcal{X}^s$. 
Now assume we have a target activity domain $\mathcal{D}^t =\{\mathcal{X}^t,P(X)^t\}$ with a marginal probability distribution $P(X)^t$ and a target activity classification task $\mathcal{T}^t$ consists of an activity label space $\mathcal{Y}^t$ where $X^t=\{x_1,\ldots,x_{n_t}\}\in \mathcal{X}^t$ and $n_t$ is number of target home sensors. In the semi-supervised setting, as we do not have all of the targets labels $Y^t$ available, we use the learning of the source domain to increase the supervised classification score in the target domain even with very few labeled data in the target domain. In our problem, the source and target domains are different but the activity classification labels are similar i.e., $\mathcal{D}^s \neq \mathcal{D}^t$, $n_s\neq n_t$, $\mathcal{X}^s \neq \mathcal{X}^t$ (also dimension $d^s\neq d^t$) but $\mathcal{T}^s \sim \mathcal{T}^s$ ($\mathcal{Y}^s \sim \mathcal{Y}^t$). 
\section{Auto-encoder Domain Adaptation}
\vspace{-0.1in}
\subsection{Deep Auto-encoder}
Auto-encoders are used to learn a compact feature representation of a certain domain with trained to reproduce an input to itself. However, auto-encoders have to map the input to a reduced dimensional representation and then reconstruct the original input from the representation. Auto-encoders consists of two parts, encoder, and decoder. Encoder maps the input feature space to a reduced dimensional representation, $\mathbf{h}=f(\mathbf{x})$ where, \textbf{x} is the input features. On the other hand, the decoder network learns to reproduce the input, $\mathbf{x} = g(\mathbf{h})$. As the encoder and the decoder network learns simultaneously, the loss function during the training is, $L(\mathbf{x},g(f(\mathbf{x})))$.
The loss function $L$ can be the mean square loss for continuous value or any other loss function. In the case of deep auto-encoder, the encoder and decoder network can be represented with a neural network consists of many CNN layers. However, the dimension of the bottleneck layer \textbf{h} must be lower than the dimension of the input feature space to extract the useful but concise feature representation. 
\par Now, we consider $N$ samples of activity window $\{\mathbf{X}^{(n)}\}^N$, where $\mathbf{X}^{(n)} \in \mathbb{R}^{n_f\times n_w}$ ; $n_f$ and $n_w$ represent the number of features and number of sensor events per window respectively. We use a single channel of the CNN network as our data windows are of a single channel. Our proposed method uses 2 layers of CNN and a single layer of Fully Connected Neural Network (FCNN) for both encoder and decoder. With the activity window $\mathbf{X}^{(n)}$:
\begin{small}
\begin{align}
    \text{CNN 1} &: \mathbf{\tilde{Z}}^{(n,c_1,1)} = \mathbf{X}^{(n)} * \mathbf{F}^{(c_1,1)}
    \text{where}, c_1 = 1,..,\mathcal{C}_1 \\
    \text{Pooling} &: \mathbf{Z}^{(n,1)} = pool(\mathbf{\tilde{Z}}^{(n,1)})\\
    \text{CNN 2} &: \mathbf{\tilde{Z}}^{(n,c_2,2)} =  \mathbf{{Z}}^{(n,1)}* \mathbf{F}^{(c_2,2)}\; \text{where}, c_2 = 1,..., \mathcal{C}_2\\
    \text{FCNN } &: h^{(n)} = \sum_{j}b_j + w_j z_j^{(n,2)} \; \text{where}, z_j^{(n,2)} \in \mathbf{\tilde{Z}}^{(n,c_2,2)}
\end{align}
\end{small}

Here, $\mathcal{C}_1$ and $\mathcal{C}_2$ are the number of filters in CNN layer 1 and layer 2 respectively. In CNN layer 1, $\mathbf{F}^{(c_1,1)}$ represents the filter tensor which convoluted with the sensor window tensor $\mathbf{X}^{(n)}$ and produce $\mathcal{C}_1$ numbers of 2D tensors. In the pooling layer, all of the tensors of filter output are stacked and reduced size according to the pooling definition and produce a 3D tensor. Similar to the first CNN layer, the output tensors of the pooling layer get convoluted by the convolution filters $\mathbf{F}^{(c_2,2)}$ and produce the tensor $\mathbf{\tilde{Z}}^{(n,c_2,2)}$. Finally, the output tensors of the second CNN layer get flattened and multiplied and added with weights and biases correspondingly in the FCNN layer. The output of the FCNN layer, $h_n$ is our reduced feature representation.

\par Similar to the encoder network, the decoder network consists of two CNN layers and single FCNN layer but in the reverse order:
\begin{scriptsize}
\begin{align}
    \text{FCNN} &: \mathbf{D}^{(n)} = \sum_{k}b_k + w_k h_k^{(n)} \; \text{where}, h_k^{(n)} \in h^{(n)}\\
    \text{CNN 2} &: \mathbf{\tilde{D}}^{(n,c_2,2)} =  \mathbf{D}^{(n)}* \mathbf{G}^{(c_2,2)}\; \text{where}, c_2 = 1,..., \mathcal{C}_2\\
    \text{Unpooling} &: \mathbf{D}^{(n,1)} = Upsampling(\mathbf{\tilde{D}}^{(n,c_2,2)})\\
    \text{CNN 1} &: \mathbf{{X}}^{(n)} = \mathbf{D}^{(n,1)} * \mathbf{G}^{(c_1,1)}\;\text{where}, c_1 = 1,...,\mathcal{C}_1 
\end{align}
\end{scriptsize}

The output of the first FCNN layer, $\mathbf{D}^{(n)}$ is reshaped to a 2D tensor and convoluted with the filter $\mathbf{G}^{(c_2,2)}$. The output 2D tensors of CNN layer 2 get stacked to a 3D tensor and upsampled in the unpooling layer. Finally, the upsampled 3D tensor get convoluted by the filter $\mathbf{G}^{(c_1,1)}$ and return the sensor window, $\mathbf{X}^{(n)}$ of the shape of original encoder input.

\subsection{Domain Adaptation with Deep Auto-encoder (AEDA)}
\begin{figure*}[t!]
\begin{center}
  \includegraphics[width=\linewidth, height=3in]{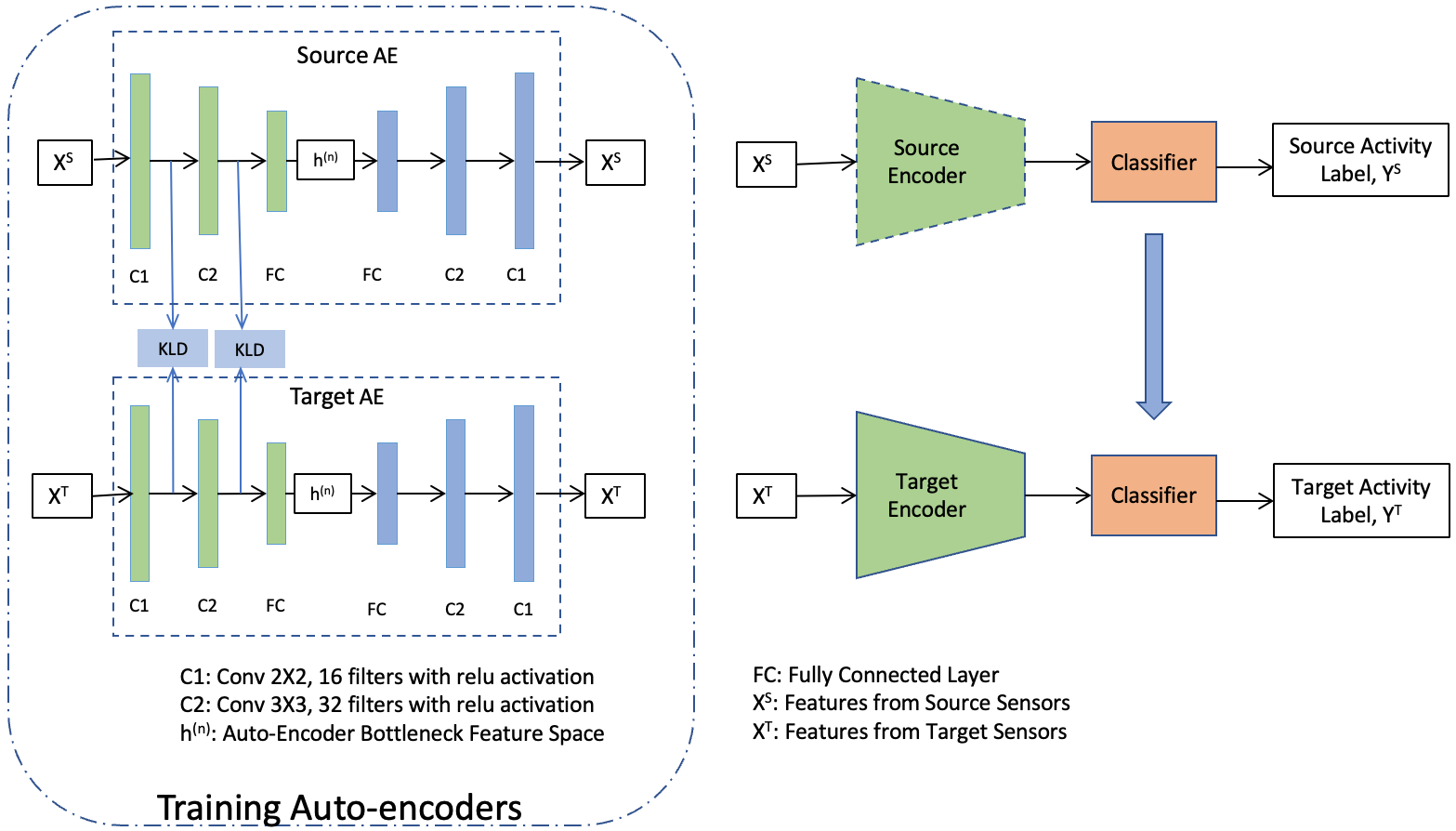}
  \caption{Proposed Auto-encoder Domain Adaptation Architeaedacture. Green and blue segments are encoder and decoder respectively. Broken line border means CNN layer weights are frozen.}
  \label{fig:aeda_diagram}
  \end{center}
\end{figure*}

\begin{figure}[h]
\begin{center}
  \includegraphics[width=\linewidth]{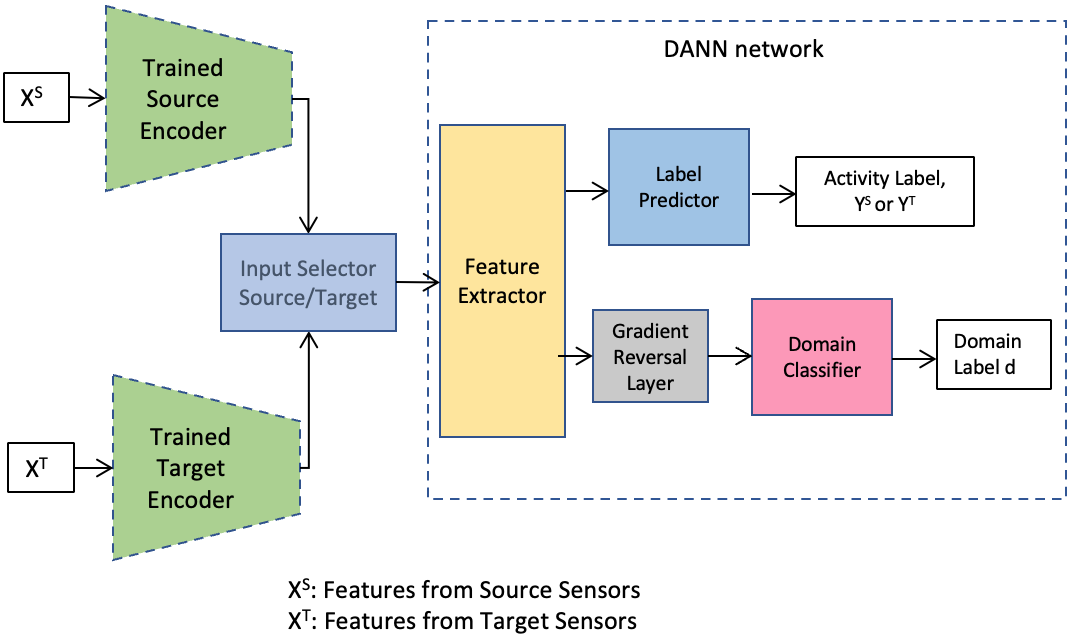}
  \caption{Enabling DANN network with Auto-encoder for Heterogeneous activity recognition. Green and blue segments are encoder and decoder respectively. Broken line border means CNN layer weights are frozen.}
  \label{fig:aedann_diagram}
  \end{center}
\end{figure}

Our proposed approach is inspired by the capability of complex feature representation of the auto-encoders. However, we use Kullback–Leibler divergence (KLD) between the corresponding CNN layer output as a loss component during the training of the target auto-encoder. This forces the target auto-encoder to map the target feature representation to a space of similar distribution to the source feature representation. The loss function we use in our proposed model is:
\vspace{-0.1in}
\begin{align}
    \text{loss} = \mathcal{L}(\mathbf{x},g(f(\mathbf{x})))+\alpha*\sum_{\text{CNN layers}}{KLD(\mathbf{\tilde{Z}}^{(n)}_{src}, \mathbf{\tilde{Z}}^{(n)}_{tgt})}
\end{align}
\vspace{-0.05in}
$\mathbf{\tilde{Z}}^{(n)}_{src}$ and $\mathbf{\tilde{Z}}^{(n)}_{tgt}$ are the CNN layer output of the corresponding source and target auto-encoder respectively. $\mathcal{L}$ is the reconstruction loss and $\alpha$ is a model parameter that was determined empirically. Figure \ref{fig:aeda_diagram} summarises our proposed architecture. The green and blue parts of the auto-encoders are encoder and decoder segments respectively. First, we train the source auto-encoder solely with the source domain features, $\{\mathcal{X}^{(s)}\}$. Then, we freeze the source encoder and add a classifier network to train with source domain features and activity label, $\mathcal{X}^s, \mathcal{Y}^s$. In the case of target AE, we train the target auto-encoder with the target domain labeled features $\{\mathcal{X}^{(s)}_l\}$. However, we use the KLD loss function along with the Mean-Squared Error (MSE) loss during the training of the target auto-encoder. Then, we append the previously learned classifier network after the encoder part of target AE and fine-tune with the target labeled data, $\{\mathcal{X}^t_l, \mathcal{Y}^t_l\}$. This enables the model to adapt to the unseen activities in the target domain. Finally, our model is ready to predict the unlabeled target domain, $\{\mathcal{X}^t_u\}$. Algorithm \ref{algo_1} shows the pseudo-code of our proposed algorithm.

\par The main insight of using KLD loss combined with regular reconstruction loss is it guides the target encoder network to achieve similar feature representation as to the source encoder step by step. If the probability distribution of the weights of a particular CNN layer of the source network is $p_s$ and the probability distribution of the weights of the corresponding target encoder CNN layer is $p_t$, the KLD of the weights of the two CNN layers is:
\vspace{-0.1in}
\begin{align}
    \begin{small}
    D_{KL}(p_s, p_t) = \sum_{i=1}^N{p_s(\mathbf{\tilde{Z}}_{si}).(\log{p_s(\mathbf{\tilde{Z}}_{si})}-\log{p_t(\mathbf{\tilde{Z}}_{ti})})}
    \end{small}
\end{align}
Where $\mathbf{\tilde{Z}}_{si}$ and $\mathbf{\tilde{Z}}_{ti}$ correspond to the $i^{th}$ output of the source and the target encoder CNN layer respectively. KL divergence represents the entropy difference between the CNN layer outputs. However, when we try to minimize the KL divergence between the layer outputs during the learning, the target weights tend to update in a way that leads to producing the output of similar distribution as possible. Applying this method layer after layer, we can achieve better domain adaptability.

\subsection{Enabling Heterogeneity with AEDA}
In this section, we explain how we can use our proposed architecture to enable heterogeneity in other algorithms which are designed to work only with homogeneous data. We implement our idea with Domain-adversarial Neural Network (DANN) \cite{ganin2016domain} architecture which is not capable of handling heterogeneous data inherently because of single input network topology. Overall architecture is presented in figure \ref{fig:aedann_diagram}.

\par Firstly, we train both of the source and the target auto-encoder just like previously mentioned \emph{AEDA} model. However, this time we freeze both of the source and the target encoder network while using the DANN network. In the native DANN architecture, only one feature extractor is used for both of the source and target input. This does not allow the different shapes of source and target data as input. For this reason, the DANN network topology only works with homogeneous source and target data. After our extension of the DANN network with two additional source and target encoder networks, now this network architecture can support two input with different tensor shape. We concatenated these two inputs into the feature extractor. As there are two input networks with the feature extractor of the DANN network, this architecture can now support heterogeneous input data. We can also accommodate heterogeneous data of different shapes by changing the input layer size of the source and the target encoder. Source and target encoder map heterogeneous feature space to a compact homogeneous feature space. In this way, we can enable virtually any homogeneous domain adaptation architecture to work with heterogeneous data. The complete procedure is summarised in Algorithm \ref{algo_2}.

\begin{algorithm}[h]
\SetAlgoLined
\begin{scriptsize}

\SetKwInOut{Input}{Input}\SetKwInOut{Output}{Output}
\SetKwRepeat{Do}{do}{while}
\Input{Labeled Source Domain, $\mathcal{D}^s = \{\mathcal{X}^s, \mathcal{Y}^s\}$, Labeled Target Domain, $\mathcal{D}^t_l = \{\mathcal{X}^t_l, \mathcal{Y}^t_l\}$, Unlabeled Target Domain, $\mathcal{D}^t_u =  \{\mathcal{X}^t_u\}$, model parameter $\alpha$, number of classifier layers $c_l$, and bottleneck space size $b$}

\Output{End to end classifier network for unlabeled target domain}
Initialize the source auto-encoder weights randomly\;
Train source AE with Source features, $\mathcal{X}^s$\;
Take only Encoder part of source AE network and append Classifier network with it\;
Freeze Encoder and randomly initialize Classifier\;
\Repeat{Test Loss converge}{
Train Encoder + Classifier network with $\mathcal{X}^s, \mathcal{Y}^s$\;
}

Initialize the target auto-encoder weights randomly\;
Set loss function as, $\text{MSE}+\alpha * \sum_{\text{CNN layers}}{KLD(\mathbf{\tilde{Z}}_{src}, \mathbf{\tilde{Z}}_{tgt})}$\;
Train target AE with labeled target features, $\mathbf{X}^t_l$\;
Take target AE and cascade with classifier network\;
Freeze target encoder part of the network\;
\Repeat{test loss converge}{
Train target encoder + classifier network with labelled target data, $\{\mathcal{X}^t_l, \mathcal{Y}^t_l\}$\;
}
Predict the label of  the target unlabelled target data, $\mathcal{D}^t_u =  \{\mathcal{X}^t_u\}$ with encoder + classifier network\;

 \caption{Auto-encoder Domain Adaptation (\emph{AEDA}) with for Heterogeneous activity recognition}
 \label{algo_1}
\end{scriptsize}
\end{algorithm}

\begin{small}

\begin{table}[h!]
\vspace{-.2in}
\caption{Overview of the CASAS datasets used}
\begin{tabular}{|c|c|c|c|c|} 
 \hline
 Domains & \multicolumn{1}{|p{1.3cm}|}{\centering Number of\\ Inhabitants} & \multicolumn{1}{|p{1.3cm}|}{\centering Number of\\ Sensors} & \multicolumn{1}{|p{1cm}|}{\centering Activity\\ Types} & Length \\ [0.2ex] 
 \hline
 hh102 & 1 & 112 & 29 & 2 months \\
 \hline
 hh113 & 1 & 123 & 32 & 2 months \\
 \hline
 hh118 & 1 & 102 & 30 & 2 months \\ [1ex] 
 \hline
 Chinook 3 & 40 & 115 & 15 & 3 months\\
 \hline

\end{tabular}
\label{dataset_overview}
\end{table}
\end{small}


\begin{algorithm}[h]
\SetAlgoLined
\begin{scriptsize}
\SetKwInOut{Input}{Input}\SetKwInOut{Output}{Output}
\SetKwRepeat{Do}{do}{while}
\Input{Labeled Source Domain, $\mathcal{D}^s = \{\mathcal{X}^s, \mathcal{Y}^s\}$, Labeled Target Domain, $\mathcal{D}^t_l = \{\mathcal{X}^t_l, \mathcal{Y}^t_l\}$, Unlabeled Target Domain, $\mathcal{D}^t_u =  \{\mathcal{X}^t_u\}$, model parameter $\alpha$, number of classifier layers $c_l$, and bottleneck space size $b$}

\Output{End to end classifier network for unlabeled target domain}
Initialize the source auto-encoder weights randomly\;
Train source AE with Source features, $\mathcal{X}^s$\;
Initialize the target auto-encoder weights randomly\;
Set loss function as, $\text{MSE}+\alpha * \sum_{\text{CNN layers}}{KLD(\mathbf{\tilde{Z}}_{src}, \mathbf{\tilde{Z}}_{tgt})}$\;
Train target AE with labeled target features, $\mathbf{X}^t_l$\;

Take only Encoder parts of both source and target AE network and use both of feature representations as to the inputs of the DANN network\;

Freeze both of the source and target Encoders and randomly initialize the DANN network\;
\Repeat{DANN network converges}{
Train Encoders + DANN network with the source data $\mathcal{D}^s = \{\mathcal{X}^s, \mathcal{Y}^s\}$ and labeled target data $\mathcal{D}^t_l = \{\mathcal{X}^t_l, \mathcal{Y}^t_l\}$\;
}

Predict the label of  the target unlabelled target data, $\mathcal{D}^t_u =  \{\mathcal{X}^t_u\}$ with encoder + DANN network\;

 \caption{Enabling DANN network with Auto-encoder for Heterogeneous activity recognition}
 \label{algo_2}
\end{scriptsize}
\end{algorithm}

\section{Experimental Setup and Evaluation}


\subsection{Datasets}
\subsubsection{CASAS \cite{cook2013casas}}
CASAS is a collection of smart-home datasets that are being widely used in activity recognition research. We use three horizon house (HH) smart-home activity datasets with single inhabitant and one multi-inhabitant smart-home activity data-set  to evaluate our transfer learning algorithms. We make a sliding window of total 10-time steps as the input feature vectors of the model. We annotate the feature windows with the mode activity of the samples. An overview of the selected datasets with the number of sensors and the number of activities is presented on table \ref{dataset_overview}. 


\subsubsection{PAMAP2 Physical Activity Monitoring Data Set \cite{reiss2012introducing}}
PAMAP2 is recorded from 9 subjects performs 18 different activities each recorded by three inertial measurement units (IMU) positioned on hand, chest and ankle respectively and a hear rate monitor sensor. We separate hand, chest and ankle data and consider as different heterogeneous domain because of different position of the sensors.

\subsection{Baseline Methods}
We implement three baseline methods and use these methods using the same dataset as the source and target to get the baseline score. The baseline score indicates the ideal condition of domain adaptation because the source and the target dataset are the same. The implemented baseline methods are: (1) Our proposed Auto-encoder Domain Adaptation (\emph{AEDA}) method; (2) HDCNN topology proposed by Khan \textit{et al.} \cite{khan2018scaling} with two CNN layers; (3) Deep Correlation Alignment (CORAL) \cite{DBLP:journals/corr/SunS16a}.

\subsection{Implementation Details}
We implement our Auto Encoder Domain Adaptation (\emph{AEDA}) network with a python based deep learning framework, Keras with Tensorflow backend. We segment the input data a sliding window of 10 samples and stride of 1 sample and feed to the model with a batch size of 128 samples per batch. However, the window size and the batch size are consistent over all of the data-sets. 
\par We use the symmetric size of the convolution layer in encoder and decoder. The first convolution layer in the encoder and the last convolution layer in the decoder have the same 16 filters with a size of $3\times3$. On the other hand, the second convolution layer in the encoder and the first convolution layer in the decoder have the same 32 filters with a size of $2\times2$. We use a max-pooling layer of $2\times2$ and $2\times1$ in the first and second layers of the encoder respectively. The reason for using different max-pooling in the different axis in the second convolution layer is the number of sensors is relatively high rather than the length of the window. Consequently, we use the Upsampling layer in the decoder concerning the max-pooling layer in the encoder. We set the model parameter $\alpha = 1\times10^{-6}$.
\par We run our \emph{AEDA} model on a server having Nvidia GTX GeForce Titan X GPU and Intel Xeon CPU (2.00GHz) processor with 12 GB of RAM. We have reported and compared the performance of different models with the common performance metric accuracy, ($\text{accuracy}=\frac{TP+TN}{TP+TN+FP+FN}$).

\begin{table*}[h!]
\addtolength{\tabcolsep}{-4pt}
\centering
\caption{Average Accuracy with 10\% Labeled Target instances in \textbf{CASAS} Single Inhabitant dataset}
\begin{tabular}{p{2.3cm}p{1.8cm}p{1.8cm}p{1.8cm}p{1.8cm}p{1.8cm}p{1.8cm}p{1.8cm}p{1.8cm}}%
    \toprule
    \bfseries Domains & \bfseries AEDA (Our)  & \bfseries AEDANN  & \bfseries MME \cite{saito2019semi} & \bfseries APE \cite{kim2020attract} & \bfseries DANN \cite{ganin2016domain} & \bfseries HDCNN  \cite{khan2018scaling} & \bfseries CORAL \cite{DBLP:journals/corr/SunS16a} & \bfseries VADA \cite{shu2018dirtt}\\
    \midrule
    hh102 to hh113 & $\mathbf{92.2 \pm 0.2}$ & $76.4 \pm 0.3$ & $87.6\pm0.3$ &$89.1\pm0.3$& $84.5 \pm 0.2$ & $73.7 \pm 0.4$ & $77.4 \pm 0.3$ & $88.4 \pm 0.2$\\
    
    hh102 to hh118 & $\mathbf{90.2 \pm 0.3}$ & $74.9 \pm 0.4$ & $88.7\pm 0.4$& $86.3\pm 0.2$ & $86.4 \pm 0.3$ & $75.8 \pm 0.2$ & $82.2 \pm 0.4$ & $84.7 \pm 0.3$\\
   
    hh113 to hh102 & $\mathbf{89.7 \pm 0.2}$ & $79.6 \pm 0.3$ &$86.3 \pm 0.4$& $87.1 \pm 0.2$ & $85.2 \pm 0.2$ & $72.8 \pm 0.2$  & $85.4 \pm 0.4$ & $85.1 \pm 0.1$\\
   
    hh113 to hh118 & $\mathbf{86.5 \pm 0.4}$ & $84.4 \pm 0.2$ & $85.3\pm0.3$ & $85.8\pm0.2$ & $82.6 \pm 0.2$ & $77.6 \pm 0.4$ & $80.1 \pm 0.2$ & $82.3 \pm 0.3$\\
   
    hh118 to hh102 & $\mathbf{88.7 \pm 0.3}$ & $83.3 \pm 0.4$ & $86.7 \pm 0.3$ & $85.6\pm0.2$ & $85.4 \pm 0.3$ & $74.4 \pm 0.4$ & $82.6 \pm 0.2$ & $84.7 \pm 0.3$\\
   
    hh118 to hh113 & $\mathbf{89.5 \pm 0.2}$ & $74.7 \pm 0.3$ & $88.1 \pm 0.4$ & $87.4 \pm 0.3$ & $87.5 \pm 0.3$ & $74.3 \pm 0.4$ & $84.3 \pm 0.5$ & $87.3 \pm 0.4$\\
    \bottomrule
   
\end{tabular}
\label{score_table_casas}
\end{table*}

\begin{table*}[h!]
\addtolength{\tabcolsep}{-4pt}
\centering
\caption{Average Accuracy with 10\% Labeled Target instances in \textbf{CASAS} Multi Inhabitant dataset}
\begin{tabular}{p{2.4cm}p{1.8cm}p{1.8cm}p{1.8cm}p{1.8cm}p{1.8cm}p{1.8cm}p{1.8cm}p{1.8cm}}%
    \toprule
    \bfseries Domains & \bfseries AEDA (Our)  & \bfseries AEDANN  & \bfseries MME \cite{saito2019semi} & \bfseries APE \cite{kim2020attract} & \bfseries DANN \cite{ganin2016domain} & \bfseries HDCNN  \cite{khan2018scaling} & \bfseries CORAL \cite{DBLP:journals/corr/SunS16a} & \bfseries VADA \cite{shu2018dirtt}\\
    \midrule
    hh102 to Chinook 3 & $\mathbf{93.1 \pm 0.3}$ & $78.5 \pm 0.2$ & $88.5\pm0.4$ &$90.2\pm0.2$& $84.6 \pm 0.3$ & $75.9 \pm 0.2$ & $79.6 \pm 0.4$ & $87.4 \pm 0.3$\\
    
    Chinook 3 to hh102 & $\mathbf{91.1 \pm 0.3}$ & $78.8 \pm 0.4$ & $89.4\pm 0.3$& $89.5\pm 0.3$ & $87.2 \pm 0.2$ & $76.5 \pm 0.2$ & $84.2 \pm 0.3$ & $86.6 \pm 0.2$\\
   
    hh113 to Chinook 3 & $\mathbf{92.8 \pm 0.3}$ & $84.7 \pm 0.4$ &$87.5 \pm 0.3$& $88.8 \pm 0.4$ & $84.6 \pm 0.3$ & $77.4 \pm 0.4$  & $86.2 \pm 0.3$ & $89.4 \pm 0.3$\\
   
    Chinook 3 to hh113 & $\mathbf{89.3 \pm 0.2}$ & $83.5 \pm 0.3$ & $86.3\pm0.4$ & $87.6\pm0.3$ & $84.6 \pm 0.3$ & $83.4 \pm 0.3$ & $82.1 \pm 0.4$ & $84.3 \pm 0.2$\\
   
    hh118 to Chinook 3 & $\mathbf{90.6 \pm 0.4}$ & $85.6 \pm 0.4$ & $87.5 \pm 0.2$ & $85.3\pm0.3$ & $87.5 \pm 0.4$ & $79.8 \pm 0.3$ & $84.6 \pm 0.2$ & $86.6 \pm 0.4$\\
   
    Chinook 3 to hh118 & $\mathbf{89.1 \pm 0.2}$ & $79.2 \pm 0.4$ & $89.3 \pm 0.2$ & $88.1 \pm 0.4$ & $89.4 \pm 0.2$ & $78.5 \pm 0.5$ & $85.4 \pm 0.4$ & $89.4 \pm 0.2$\\
    \bottomrule
   
\end{tabular}
\label{score_table_casas_multi}
\end{table*}

\begin{table*}[h!]
\addtolength{\tabcolsep}{-4pt}
\centering
\caption{Average Accuracy with 10\% Labeled Target instances in \textbf{PAMAP2} dataset}
\begin{tabular}{p{2.3cm}p{1.8cm}p{1.8cm}p{1.8cm}p{1.8cm}p{1.8cm}p{1.8cm}p{1.8cm}p{1.8cm}}%
    \toprule
    \bfseries Domains & \bfseries AEDA (Our)  & \bfseries AEDANN & \bfseries MME \cite{saito2019semi} & \bfseries APE \cite{kim2020attract} & \bfseries DANN \cite{ganin2016domain} & \bfseries HDCNN  \cite{khan2018scaling} & \bfseries CORAL \cite{DBLP:journals/corr/SunS16a} & \bfseries VADA \cite{shu2018dirtt} \\
    \midrule
    hand to chest & $\mathbf{94.2 \pm 0.4}$ & $86.3 \pm 0.2$ & $92.3 \pm 0.4$ & $93.4 \pm 0.3$ & $90.4 \pm 0.3$ & $82.6 \pm 0.3$ & $85.3 \pm 0.3$ & $93.3 \pm 0.3$\\
    
    hand to ankle & $\mathbf{92.7 \pm 0.2}$ & $87.4 \pm 0.3$ & $90.2 \pm 0.2$ & $89.6 \pm 0.5$ & $91.8 \pm 0.4$ & $84.9 \pm 0.4$ & $87.6 \pm 0.2$ & $91.9 \pm 0.4$\\
   
    chest to ankle & $\mathbf{91.1 \pm 0.3}$ & $85.5 \pm 0.4$ & $89.4 \pm 0.3$ & $88.8 \pm 0.4$ & $88.4 \pm 0.5$ & $83.5 \pm 0.3$  & $89.4 \pm 0.1$ & $90.5 \pm 0.5$\\
   
    chest to hand & $90.4 \pm 0.5$ & $83.6 \pm 0.3$ & $85.6 \pm 0.4$ & $86.4 \pm 0.4$ & $89.4 \pm 0.2$ & $88.7 \pm 0.2$ & $\mathbf{91.1 \pm 0.3}$ & $85.3 \pm 0.2$\\
   
    ankle to hand & $\mathbf{93.1 \pm 0.2}$ & $89.8 \pm 0.5$ & $91.4 \pm 0.3$ & $91.6 \pm 0.4$ & $92.9 \pm 0.4$ & $86.3 \pm 0.1$ & $92.4 \pm 0.4$ & $91.8 \pm 0.4$\\
   
    ankle to chest & $\mathbf{91.5 \pm 0.3}$ & $86.7 \pm 0.3$ & $89.4 \pm 0.5$ & $90.6 \pm 0.4$ & $88.7 \pm 0.2$ & $81.1 \pm 0.2$ & $90.3 \pm 0.4$ & $89.8 \pm 0.2$\\
    \bottomrule
   
\end{tabular}
\label{score_table_pamap2}
\end{table*}

\subsection{Results}
\begin{figure*}[h]
  \includegraphics[width=\linewidth, height = 2.5 in]{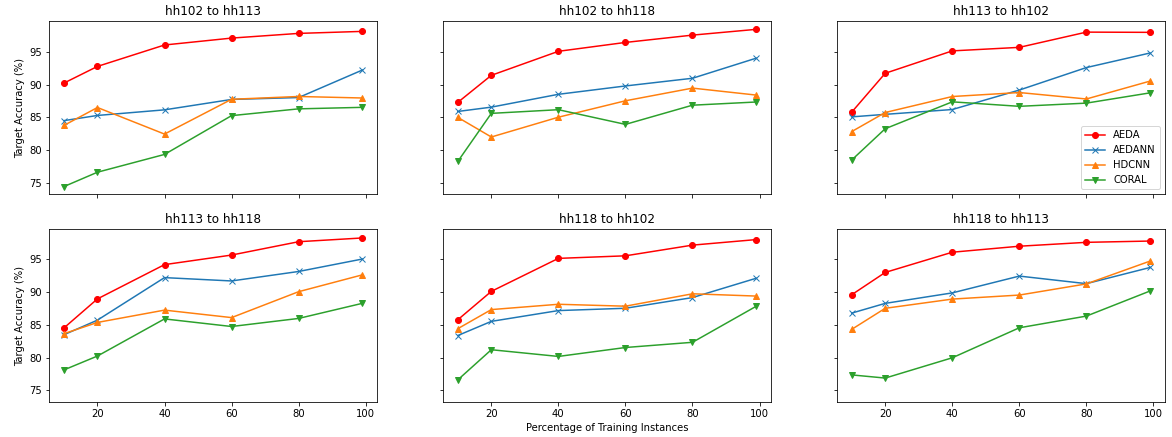}
  \caption{Accuracy comparison with different percentage of labeled training data in CASAS dataset}
  \label{fig:accuracy}
  \vspace{-0.2in}
\end{figure*}

We evaluate the performance of our proposed framework mainly on two scenarios. i) The baseline method where the source and the target datasets are the same, and ii) all of the combinations of three datasets both from CASAS and PAMAP2 separately. Additionally, we also evaluate our algorithm in the following cases:
\begin{itemize}
    \item Accuracy with a varying fraction of labeled data.
    \item Effect of features and class diversity.
    \item Accuracy on unseen activities.
\end{itemize}
\subsubsection{Baseline Accuracy}
\begin{figure}
  \includegraphics[width=\linewidth, height=4cm]{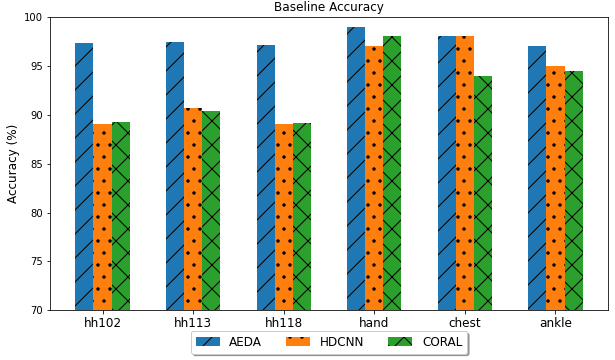}
  \caption{Baseline accuracy comparison of the models}
  \label{fig:baseline}
  \vspace{-0.1in}
\end{figure}

We established baseline accuracy for all three models to get the data quality. The baseline methods use the same domain as both the source and the target domain of the models. As no transfer of learning is happening in this case, this baseline accuracy shows the maximum achievable accuracy of a particular model. Figure \ref{fig:baseline} shows the performance comparison among the models which clearly illustrates that our proposed model \emph{AEDA} outperforms all other methods in all 6 domains in all two datasets. Both HDCNN and CORAL perform very similarly in all three domains of CASAS datasets with about $\approx 90\%$ accuracy. Performance of both of the models is about $96\thicksim98\%$ in case of the three domains of PAMAP2 dataset. Finally, our proposed Auto-encoder Domain Adaptation (\emph{AEDA}) shows about $97\thicksim99\%$ accuracy with all of the six domains from both of the datasets which is a significant improvement over HDCNN and CORAL model.

\subsubsection{Domain Adaptation Accuracy}
We compared the performance of our proposed \emph{AEDA} model and auto-encoder enabled DANN model for heterogeneous application with the following six bench-marking domain adaptation frameworks: Minimax Entropy (MME)\cite{saito2019semi}, Attract, Perturb, and Explore (APE)\cite{kim2020attract}, Domain  Adversarial  Neural  Network (DANN) \cite{ganin2016domain}, Heterogeneous Deep Convolutional Neural Network (HDCNN) \cite{khan2018scaling}, Deep  Correlation  Alignment  (Deep-CORAL) \cite{DBLP:journals/corr/SunS16a} and Virtual Adversarial Domain Adaptation (VADA) \cite{shu2018dirtt}.
We implement all of these algorithms using deep domain adaptation framework SALAD \cite{schneider2018salad}. We test the algorithms with all of the combinations of source and target domains with CASAS and PAMAP2 datasets. Different domains in the CASAS dataset present different smart-home with completely different sensor setup and subject. This presents the sensor heterogeneity. Besides, in the PAMAP2 dataset, different domains represent the position of the IMU sensor, i.e, hand, chest and ankle which shows position heterogeneity. The use of data with different sensor diversity and class diversity in source and target ensures the robustness of our proposed architecture. The performance comparisons with the four benchmark models with a different combination of source and target dataset are presented in table \ref{score_table_casas} and table \ref{score_table_pamap2} for CASAS and PAMAP2 dataset respectively. As our algorithm is a semi-supervised learning method, we use 10\% of the target data as labeled and the rest of the data as unlabeled. The performance of our proposed method outperforms all other models with a significant margin in almost all combinations of source and target domains. Our \emph{AEDA} model achieves $85\thicksim 90$\% of accuracy with different sources and target domains in CASAS dataset and about $90\thicksim94\%$ in PAMAP2 dataset.

\subsubsection{Accuracy with Varying Fraction of Labeled Data}
We study the performance of our proposed architecture with a different fraction of labeled data in the target dataset. The primary questions of our interest are: (i) what percentage of labeled data can contribute to a significant improvement of the performance? (ii) Is our proposed model really justified in the scenario of a low fraction of labeled data? We compare the accuracy changes with the changes of labeled data fraction with different baseline algorithms and use all the six combinations of CASAS dataset in figure \ref{fig:accuracy}.

\par In order to investigate the performance improvement with the change of the labeled data fraction, we train the model with the source data first. Then, we divide the target dataset into a labeled and an unlabeled segment randomly. Consequently, we fine-tune the model with the labeled fraction of the data. Finally, we evaluate the performance of the model on the unlabeled segment of the data. 

\par From figure \ref{fig:accuracy} we find that our proposed architecture outperforms all other baseline models in all combinations of source and target dataset. Our proposed AEDA model achieves near-maximum accuracy only after $\approx 40\%$ of labeled training instances. This is really important as we can achieve near supervised accuracy with a fraction of labeled data. This advantage of our proposed algorithm can be used in scenarios where labeling data is expensive. However, a more interesting fact is: the performance of our proposed \emph{AEDA} model is more stable with respect to both HDCNN and CORAL models as the percentage of labeled training instances changes. Moreover, the performance is also consistent over the different source and target domains.

\subsubsection{Different Features and Classes Diversity on Target Dataset}
We also study our proposed algorithm in scenarios of different features and class diversity on the target domains in CASAS dataset. We determine the diversity of features and classes based on new classes and features appear on the target domain other than the features and classes already learned on the source domain.  

\begin{table}[h!]
\caption{Sensor and Class Diversity on Target Domain in Casas dataset}
\begin{tabular}{|c|c|c|c|c|} 
 \hline
 Domain 
 & \multicolumn{1}{|p{1cm}|}{\centering Unseen \\ Sensors}
 & \multicolumn{1}{|p{1cm}|}{\centering Sensors \\ Diversity}
 & \multicolumn{1}{|p{1cm}|}{\centering Unseen \\ Class} 
 & \multicolumn{1}{|p{1cm}|}{\centering Class \\ Diversity} \\ [0.5ex]
 \hline
 hh102 to hh113 & 21 & High & 4 & High \\ 
 \hline
 hh102 to hh118 & 27 & High & 5 & High \\
 \hline
 hh113 to hh102 & 10 & Low & 1 & Low \\  
  \hline
 hh113 to hh118 & 14 & Low & 2 & Low \\  
  \hline
 hh118 to hh102 & 37 & High & 4 & High \\  
  \hline
 hh118 to hh113 & 35 & High & 4 & High \\  
 \hline
\end{tabular}
\label{diversity_table}
\end{table}

\begin{figure}
  \includegraphics[width=\linewidth, height=3.5cm]{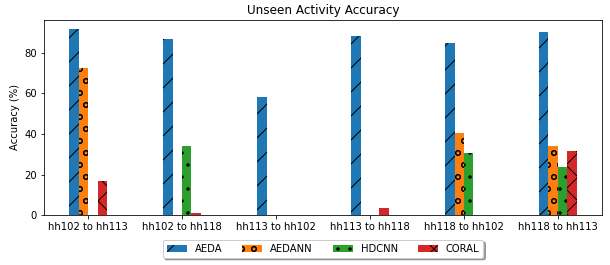}
  \caption{Accuracy score on unseen activities on CASAS dataset}
  \label{fig:unseen}
  \vspace{-0.1in}
\end{figure}

\par Table \ref{diversity_table} shows the sensor and class diversity in different domains of source and target data based on our criteria. Our criteria for high sensor diversity is having more than 15 unseen sensors in the target dataset. Similarly, more than 3 unseen class in the target dataset are considered as high-class diversity. According to table \ref{diversity_table}, `hh113 to hh102' and `hh113 to hh118' are considered as low sensors and low-class diversity and the rest of the domains are considered as high sensors and high-class diversity. From figure \ref{fig:accuracy}, it is evident that low sensors and class diversity domains result in higher stability on accuracy. On the other hand, domains having high sensors and class diversity show lower stability of accuracy over the different fraction of target instances. 

\subsubsection{Accuracy on Unseen Activities}
We compare the performance on unseen activities that are newly appeared in the target dataset. 
Figure \ref{fig:unseen} shows the accuracy performance of unseen activities over different combinations of source and target dataset. It is clear that our proposed \emph{AEDA} algorithm outperforms all other baseline algorithms overall of the domains. In most of the cases, \emph{AEDA} achieves over $\approx 90 \%$ accuracy where HDCNN, AEDANN and CORAL show very poor performances. This shows the strength of our proposed algorithm over other baseline models in the scenario of heterogeneous domain adaptation.
\par The main reason for superior performance on unseen data is in the generalization capability over the fundamental components of certain activity. During the mapping to a common feature space in the bottleneck layer of the auto-encoders, similar components of different activities get mapped in nearby regions. As a result when some fundamental components of an unseen activity get matched with and seen activity, the model can easily classify those unseen activities.

\begin{figure}
  \includegraphics[width=\linewidth, height=4.5cm]{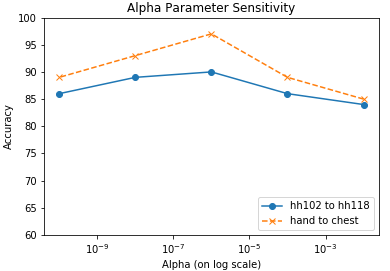}
  \caption{Parameter Sensitivity Analysis on Alpha parameter.}
  \label{fig:param_sens}
  \vspace{-0.2in}
\end{figure}

\subsubsection{Parameter Sensitivity Analysis} We choose the optimal value of the model parameter $\alpha$ by analyzing the effect of different values of $\alpha$ on the performance of our model. Figure \ref{fig:param_sens} shows the accuracy score over different values of alpha on two different domain setups from CASAS and PAMAP2 datasets. The accuracy score does not change drastically on the change of alpha which shows the robustness of our model in terms of the change of parameters. We choose the optimal value of alpha as $\alpha = 1\times10^{-6}$ from this analysis.

\subsubsection{Ablation Study}
We perform ablation study with CASAS and PAMAP2 datasets to understand the performance change with different loss functions. We perform the experiments by removing KLD loss component between the bottleneck layer. The significant drop of performance in both CASAS and PAMAP2 datasets in table \ref{tab:ablation} shows the importance of KLD loss function in our model architecture.
\begin{table}[h]
\captionsetup{font=small}
    \centering
      \caption{Effect of KLD loss function and simultaneous learning on CASAS and PAMAP2 data}%
      \begin{tabular}{c|c|c}
        \hline
        Approach &  CASAS &  PAMAP2 \\
        \hline
        w/o KLD Loss (w CrossEntropy Loss) & $78.4\pm0.6$ & $80.5\pm0.5$ \\
        \hline
        Our Method & $89.3\pm0.3$& $91.6\pm0.2$ \\
        \hline
    \end{tabular}
    \vspace{-0.1cm}

  \label{tab:ablation}
  \vspace{-0.1in}
\end{table}

\section{Discussion and Future Work}
Unsupervised domain adaptation (UDA) methods improve generalization on unlabeled target data by aligning distributions but can fail to learn discriminative class boundaries on target domains. However, in Semi-supervised Domain Adaptation (SDA) setting where a few target labels are available, existing methods often do not improve performance relative to just training on the labeled source and target examples, and can even make it worse by imposing negative learning. Moreover, the presence of heterogeneity between the domains can, sometimes, make it impossible to work every other domain adaptation method due to the presence of diversity both in feature and label space. Our proposed \emph{AEDA} offers a novel method that can ease the domain adaptation in heterogeneous settings both for the feature and class diversity. Our proposed \emph{AEDA} is powerful enough to enable the existing domain adaptation framework to work in a heterogeneous setting.

Though our result shows a significantly improved result with our proposed \emph{AEDA} method, there are few limitations of the framework which we aim to address in the future. First, our method enables the DANN architecture for heterogeneous activity data but the performance is poor compared to our vanilla \emph{AEDA} architecture. This may be because of the reduced capability of the adversarial part of DANN on the representation space which we aim to investigate in the future. We implement and evaluate the performance of all of our algorithms on 12 different combinations of 6 different heterogeneous domains in two publicly available activity datasets where ambient and wearable sensors have been used separately. 

\section{Conclusion}
Many researchers proposed Semi-supervised domain adaptation (SSDA) with applicability in computer vision and Natural Language Processing (NLP) showing few problem specific successes which are mostly dependent on homogeneity between source and target domains. We develop \emph{AEDA} on strong basis of the gaps existed between the applicability of homogeneous deep domain adaptation in heterogeneous setting, especially, on heterogeneous smart home sensors (ambient and wearable) environment. Our experimental result shows that \emph{AEDA} outperforms all other existing baseline algorithms in different scenarios of heterogeneity of the activity recognition data such as low and high heterogeneity in feature and class spaces. Our method also outperforms other algorithms in terms of unseen activities in the target domain. Further, we believe that, \emph{AEDA}, first of its kind framework, that has the capability to enable any homogeneous deep learning algorithm (supervised, unsupervised, semi-supervised, self-supervised) to handle heterogeneous domains that signifies the utmost importance of this innovation.

\section{Acknowledgments}
This work is partially supported by NSF’s Smart \& Connected Community award \#2230180


\bibliographystyle{IEEEtran}
\bibliography{reference}

\begin{thebibliography}{10}
\providecommand{\url}[1]{#1}
\csname url@samestyle\endcsname
\providecommand{\newblock}{\relax}
\providecommand{\bibinfo}[2]{#2}
\providecommand{\BIBentrySTDinterwordspacing}{\spaceskip=0pt\relax}
\providecommand{\BIBentryALTinterwordstretchfactor}{4}
\providecommand{\BIBentryALTinterwordspacing}{\spaceskip=\fontdimen2\font plus
\BIBentryALTinterwordstretchfactor\fontdimen3\font minus
  \fontdimen4\font\relax}
\providecommand{\BIBforeignlanguage}[2]{{%
\expandafter\ifx\csname l@#1\endcsname\relax
\typeout{** WARNING: IEEEtran.bst: No hyphenation pattern has been}%
\typeout{** loaded for the language `#1'. Using the pattern for}%
\typeout{** the default language instead.}%
\else
\language=\csname l@#1\endcsname
\fi
#2}}
\providecommand{\BIBdecl}{\relax}
\BIBdecl

\bibitem{lara2012survey}
O.~D. Lara and et. al., ``A survey on human activity recognition using wearable
  sensors,'' \emph{IEEE communications surveys \& tutorials}, vol.~15, no.~3,
  2012.

\bibitem{DBLP:journals/corr/SunS16a}
B.~Sun and et. al., ``Deep {CORAL:} correlation alignment for deep domain
  adaptation,'' \emph{CoRR}, vol. abs/1607.01719, 2016.

\bibitem{tzeng2017adversarial}
E.~Tzeng and et. al., ``Adversarial discriminative domain adaptation,'' in
  \emph{CVPR}, 2017.

\bibitem{ganin2016domain}
Y.~Ganin and et. al., ``Domain-adversarial training of neural networks,''
  vol.~17, no.~1, 2016.

\bibitem{wang18}
J.~Wang and et. al., ``Stratified transfer learning for cross-domain activity
  recognition,'' in \emph{Percom}, 2018.

\bibitem{wyatt2005unsupervised}
D.~Wyatt and et. al., ``Unsupervised activity recognition using automatically
  mined common sense,'' in \emph{AAAI}, vol.~5, 2005.

\bibitem{attal2015physical}
F.~Attal and et. al., ``Physical human activity recognition using wearable
  sensors,'' \emph{Sensors}, vol.~15, no.~12, 2015.

\bibitem{glorot2011domain}
X.~Glorot, A.~Bordes, and Y.~Bengio, ``Domain adaptation for large-scale
  sentiment classification: A deep learning approach,'' 2011.

\bibitem{long2016unsupervised}
M.~Long and et. al., ``Unsupervised domain adaptation with residual transfer
  networks,'' in \emph{Advances in neural information processing systems},
  2016.

\bibitem{chen2012marginalized}
M.~Chen and et. al., ``Marginalized denoising autoencoders for domain
  adaptation,'' \emph{arXiv preprint arXiv:1206.4683}, 2012.

\bibitem{kan2015bi}
M.~Kan, S.~Shan, and X.~Chen, ``Bi-shifting auto-encoder for unsupervised
  domain adaptation,'' in \emph{CVPR}, 2015.

\bibitem{ming18}
M.~Long and et. al., ``Conditional adversarial domain adaptation,'' 2014.

\bibitem{kuniaki18}
K.~Saito and et. al., ``Adversarial dropout regularization,'' 2018.

\bibitem{daume2010frustratingly}
H.~Daum{\'e}~III and et. al., ``Frustratingly easy semi-supervised domain
  adaptation,'' in \emph{Workshop on Domain Adaptation for Natural Language
  Processing}.\hskip 1em plus 0.5em minus 0.4em\relax ACL, 2010.

\bibitem{xiao2014feature}
M.~Xiao and Y.~Guo, ``Feature space independent semi-supervised domain
  adaptation via kernel matching,'' \emph{IEEE transactions on pattern analysis
  and machine intelligence}, vol.~37, no.~1, 2014.

\bibitem{ao2017fast}
S.~Ao, X.~Li, and C.~X. Ling, ``Fast generalized distillation for
  semi-supervised domain adaptation,'' in \emph{Thirty-First AAAI Conference on
  Artificial Intelligence}, 2017.

\bibitem{li2013learning}
W.~Li, L.~Duan, D.~Xu, and I.~W. Tsang, ``Learning with augmented features for
  supervised and semi-supervised heterogeneous domain adaptation,'' \emph{IEEE
  transactions on pattern analysis and machine intelligence}, vol.~36, no.~6,
  2013.

\bibitem{wang2011heterogeneous}
C.~Wang and S.~Mahadevan, ``Heterogeneous domain adaptation using manifold
  alignment,'' in \emph{Twenty-second international joint conference on
  artificial intelligence}, 2011.

\bibitem{hubert2016learning}
Y.-H. Hubert~Tsai, Y.-R. Yeh, and Y.-C. Frank~Wang, ``Learning cross-domain
  landmarks for heterogeneous domain adaptation,'' in \emph{CVPR}, 2016.

\bibitem{yan2017learning}
Y.~Yan and et. al., ``Learning discriminative correlation subspace for
  heterogeneous domain adaptation.'' in \emph{IJCAI}, 2017.

\bibitem{khan2018scaling}
M.~A. A.~H. Khan, N.~Roy, and A.~Misra, ``Scaling human activity recognition
  via deep learning-based domain adaptation,'' in \emph{2018 IEEE International
  Conference on Pervasive Computing and Communications (PerCom)}.\hskip 1em
  plus 0.5em minus 0.4em\relax IEEE, 2018.

\bibitem{cook2013casas}
D.~J. Cook, A.~S. Crandall, B.~L. Thomas, and N.~C. Krishnan, ``Casas: A smart
  home in a box,'' \emph{Computer}, vol.~46, no.~7, 2013.

\bibitem{reiss2012introducing}
A.~Reiss and D.~Stricker, ``Introducing a new benchmarked dataset for activity
  monitoring,'' in \emph{2012 16th International Symposium on Wearable
  Computers}.\hskip 1em plus 0.5em minus 0.4em\relax IEEE, 2012.

\bibitem{saito2019semi}
K.~Saito, D.~Kim, S.~Sclaroff, T.~Darrell, and K.~Saenko, ``Semi-supervised
  domain adaptation via minimax entropy,'' in \emph{ICCV}, 2019.

\bibitem{kim2020attract}
T.~Kim and C.~Kim, ``Attract, perturb, and explore: Learning a feature
  alignment network for semi-supervised domain adaptation,'' in \emph{European
  Conference on Computer Vision}.\hskip 1em plus 0.5em minus 0.4em\relax
  Springer, 2020.

\bibitem{shu2018dirtt}
R.~Shu, H.~H. Bui, H.~Narui, and S.~Ermon, ``A dirt-t approach to unsupervised
  domain adaptation,'' 2018.

\bibitem{schneider2018salad}
S.~Schneider, A.~S. Ecker, J.~H. Macke, and M.~Bethge, ``Salad: A toolbox for
  semi-supervised adaptive learning across domains,'' 2018.

\end{thebibliography}

\newpage

 




\vfill

\end{document}